\documentclass{article}

\usepackage{microtype}
\usepackage{graphicx}
\usepackage{subfigure}
\usepackage{booktabs} %
\usepackage{amsmath}
\usepackage{amssymb}
\usepackage{bbm}
\usepackage{amsfonts}
\usepackage{todonotes}
\usepackage{algorithm}
\usepackage{todonotes} %
\usepackage{float}
\usepackage{soul}

\usepackage{mathtools}

\DeclareMathOperator*{\argmax}{arg\,max}

\newcommand{\D}{\mathcal{D}}
\newcommand{\U}{\mathcal{U}}
\newcommand{\St}{\mathcal{S}}
\newcommand{\A}{\mathcal{A}}
\newcommand{\R}{\mathbb{R}}
\newcommand{\E}{\mathop{\mathbb{E}}}
\newcommand{\defeq}{\vcentcolon=}

\usepackage{hyperref}
\usepackage{cleveref}

\usepackage[accepted]{icml2021}

\icmltitlerunning{Offline Reinforcement Learning with Fisher Divergence Critic Regularization}

\begin{document}

\twocolumn[
\icmltitle{Offline Reinforcement Learning \\ with Fisher Divergence Critic Regularization}

\begin{icmlauthorlist}
\icmlauthor{Ilya Kostrikov}{nyu,google}
\icmlauthor{Jonathan Tompson}{google}
\icmlauthor{Rob Fergus}{nyu,deepmind}
\icmlauthor{Ofir Nachum}{google}
\end{icmlauthorlist}

\icmlaffiliation{nyu}{New York University, USA}
\icmlaffiliation{google}{Google Research, USA}
\icmlaffiliation{deepmind}{Google DeepMind, USA}

\icmlcorrespondingauthor{Ilya Kostrikov}{ikostrkov@gmail.com}

\icmlkeywords{Machine Learning, ICML}

\vskip 0.3in
]

\printAffiliationsAndNotice{}  %

\begin{abstract}
Many modern approaches to offline Reinforcement Learning (RL) utilize \emph{behavior regularization}, typically augmenting a model-free actor critic algorithm with a penalty measuring divergence of the policy from the offline data. 
In this work, we propose an alternative approach to encouraging the learned policy to stay close to the data, namely parameterizing the critic as the $\log$-behavior-policy, which generated the offline data, plus a state-action value offset term, which can be learned using a neural network.
Behavior regularization then corresponds to an appropriate regularizer on the offset term.
We propose using a gradient penalty regularizer for the offset term and demonstrate its equivalence to Fisher divergence regularization, suggesting connections to the score matching and generative energy-based model literature.
We thus term our resulting algorithm Fisher-BRC (Behavior Regularized Critic).
On standard offline RL benchmarks, Fisher-BRC achieves both improved performance and faster convergence over existing state-of-the-art methods.  \footnote{Code to reproduce our results is available at \url{https://github.com/google-research/google-research/tree/master/fisher_brc}.}
\end{abstract}

\section{Introduction}
Reinforcement learning (RL) describes the field of machine learning concerned with learning a policy to solve a task through stochastic trial-and-error experience in an environment.
The default setting typically assumed in RL is of \emph{online} access to the environment; i.e., the learned policy may collect new trial-and-error experience directly from the environment.
However, in many practical scenarios, where deploying a new policy to interact with the live environment is expensive or associated with risks or safety concerns~\citep{thomas2015safe}, it is more common to have only \emph{offline} access to the environment. 
That is, the trial-and-error experience available for learning a task-solving policy is a static, offline dataset of experience collected by some other \emph{behavior} policy.
This setting is known as offline RL and has attracted a significant amount of interest in recent years~\citep{levine2020offline,lange2012batch}.

Many of the recent approaches to offline RL utilize some form of \emph{behavior regularization}, in which the learned policy is compelled to stay close to the data-generating behavior policy~\citep{wu2019behavior}.
For example, in model-free actor critic algorithms, behavior regularization is typically done by augmenting the actor loss with a penalty measuring the divergence of the learned policy from the behavior policy~\citep{jaques2019way,kumar19bear,wu2019behavior}, reminiscent of KL-control methods in related literature~\citep{kappen2012optimal,jaques2017sequence}. 
While straightforward, a disadvantage of this approach is that it does little to regularize the critic itself, and thus, it is common for the critic to take on wildly extrapolated values on actions unseen in the training data, which may dominate any behavior regularization applied to the actor, as we demonstrate in this work.

In this work, we propose an alternative approach to encouraging the learned policy to stay close to the offline data. Focusing on the critic, we propose parameterizing the critic values as the logits of the behavior policy plus an additional offset term. When the actor (the learned policy) is then trained to choose actions which maximize the critic value, it will be compelled to stay close to the behavior policy as long as the offset term is suitably `small'.
Thus, behavior regularization in this setting corresponds to augmenting the standard Bellman error critic loss with an appropriate regularization on the offset term.

What should this regularization term be? 
After noting that, in continuous control, the offset term's effect on the learned policy is via the gradients of the offset with respect to actions, we propose regularizing the offset term with a gradient penalty.
While this may appear heuristic at first, we present mathematical derivations establishing a connection between this gradient penalty and the \emph{Fisher divergence}, which appears in the score matching and energy-based generative model literature~\citep{lyu2012interpretation,bao2020variational}, interpreting the critic values as the energy function of a Boltzmann distribution.

We thus term our newly proposed actor critic algorithm \emph{Fisher-BRC} (behavior regularized critic). 
To aid conceptual understanding of Fisher-BRC, we analyze its training dynamics in a simple toy setting, highlighting the advantage of its implicit Fisher divergence regularization as opposed to the more explicit divergence penalties imposed by alternative offline RL methods.
We then present an extensive evaluation of Fisher-BRC on standard offline RL benchmarks. 
We find that Fisher-BRC yields state-of-the-art performance compared to a variety of existing model-free and model-based RL methods. 
We further show that while Fisher-BRC learns better policies, it is also much more computationally efficient than more sophisticated offline RL algorithms.
Overall, Fisher-BRC presents a new approach to behavior regularization in offline RL with compelling practical benefits.

\section{Related Work}

Our work adds to the rich literature on behavior regularization methods in offline RL~\citep{wu2019behavior}, which propose a number of regularizations in RL training that compel the learned policy to stay close to the offline data.
These regularizers have appeared as divergence penalties~\citep{kumar19bear,jaques2019way,wu2019behavior}, implicitly through appropriate network initializations~\citep{matsushima2020deployment}, or more explicitly through careful parameterization of the policy~\citep{fujimoto2019off}.
Another way to apply behavior regularizers is via modification of the critic learning objective as in \cite{nachum2019algaedice, kumar2020conservative}. 
Our work is unique in applying behavior regularization through a parameterization of the critic, and empirically, we find this to yield much better performance.

In this work, we demonstrate that the specific parameterization we employ has connections to Fisher divergence regularization, establishing connections with energy-based models.
Prior methods have proposed using energy-based models for policies to express more multi-modal distributions~\citep{haarnoja2017reinforcement,pmlr-v24-heess12a}.
Meanwhile, while our work focuses on reinforcement learning -- i.e., learning a return-maximizing policy -- other works have established connections between energy-based models and inverse RL or imitation learning~\citep{finn2016connection}.

In practice, our regularization reduces to a gradient penalty applied to the offset term in the critic. 
Gradient penalties have appeared in previous work, arguably first popularized in machine learning by Wasserstein generative-adversarial networks~\citep{arjovsky2017wasserstein}, but also used in imitation learning~\citep{kostrikov2018discriminatoractorcritic}, cross-domain disentanglement~\citep{gonzalezgarcia2018imagetoimage}, and uncertainty estimation~\citep{vanamersfoort2020uncertainty}.

\section{Background}
We introduce the notation and assumptions used in this paper, as well as provide an in-depth review of the methods most closely related to ours. 

\subsection{Reinforcement Learning}
In this work, we consider environments that can be represented as a Markov Decision Process (MDP) defined by a tuple $(\St, \A, p_0, p, r, \gamma)$, where $\St$ is a state space, $\A$ is an action space, $p_0(s)$ is a distribution of initial states, $p(s'|s,a)$ is a stochastic dynamics model, $r: \St \times \A \rightarrow \R$ is a reward function and $\gamma \in [0, 1)$ is a discount. 
We restrict our work to continuous action spaces; i.e., $\A\subset \R^{d}$ for some $d$.
The goal of reinforcement learning is to find a policy $\pi(a|s)$ that maximizes the cumulative discounted returns $\E_\pi[\sum_{t=0}^\infty\gamma^t r(s_t, a_t) | \allowbreak s_0 \sim p_0(\cdot), \allowbreak  a_t \sim \pi(\cdot|s_t), \allowbreak s_{t+1} \sim p(\cdot|s_t, a_t) ]$. In online reinforcement learning, it is usually assumed that an agent learns based on experience generated by the agent interacting with the learning environment. %

\subsection{Offline Reinforcement Learning}
In this work, we focus on the offline setting, in which the agent cannot generate new experience data and learns based on a provided dataset $\D$ of $(s, a, r, s')$ tuples generated by some other policy interacting with the environment. 
We call the policy that generated this dataset the \emph{behavior} policy and denote it as $\mu$.

Offline datasets usually do not provide complete state-action coverage. That is, the set $\{(s,a) ~|~ (s,a,r,s')\in\D\}$ is typically a small subset of the full space $\St\times\A$. 
Standard reinforcement learning methods such as SAC~\citep{haarnoja2019soft} or DDPG~\citep{lillicrap2019continuous} suffer when applied to these datasets due to extrapolation errors \cite{fujimoto2019off, kumar19bear}. 

Behavior regularization is a prominent approach to offline reinforcement learning that aims to address this problem by using appropriate regularizers to compel the learned policy to stay close to the data. There are two common ways to incorporate behavior regularization into the actor-critic framework -- via policy regularization or via a critic penalty. Since our own approach is related to both techniques, in the following sections we describe the two approaches as well as problems associated with them. 

\subsection{Policy Regularization}
Policy regularization can be imposed either during critic or policy learning. First, we describe a family of approaches based on applying behavior constraints to training policies. These constraints can be applied in a hard fashion, by restricting the policy action space to the actions seen in the offline dataset as in BCQ~\cite{fujimoto2019off}:
\vspace{-2mm}
\begin{align*}
\begin{split}
\pi(s) \defeq \argmax_{a_i + \xi_\phi(s,a_i,\Phi)} Q_\theta(s, a_i + \xi_\phi(s,a_i,\Phi)),\\
\{a_i\sim \mu(\cdot|s)\}_{i=1}^n
\end{split}
\end{align*}
where $\pi(s)$ is a deterministic policy, $\xi_\phi(s,a,\Phi)$ is a perturbation model on actions constrained to the interval $[-\Phi, \Phi]$, $\mu(a|s)$ is a behavioral policy contracted by fitting a density model to the offline dataset. The hyperparameter $n$ controls the number of sampled actions. In other words, one samples $n$ perturbed actions from the behavior policy and chooses from the action with the largest critic-approximated value. One of the limitations of this approach is that a large number of sampled actions might be required for competitive performance ~\cite{ghasemipour2021emaq}. %

Using a divergence penalty is an alternative approach to policy regularization. Instead of having hard constraints on the training policy, one can regularize the policy with an appropriately chosen probability divergence such as the KL-divergence~\citep{wu2019behavior,jaques2019way}:
\vspace{-2mm}
\begin{equation}
\label{eq:kl_control}
    \hspace{-4mm}\max_{\pi} \hspace{-1mm}\E_{s \sim \D}\left[\E_{a \sim \pi(\cdot|s)} [Q_\theta(s,a)] - \alpha D_{KL}(\pi(\cdot|s)\|\mu(\cdot|s)) \right].\hspace{-2mm}
\end{equation}

Although these approaches demonstrate impressive performance on some tasks, they share a common problem. The Q-function, $Q_\theta$, learned via standard TD-error minimization on $\D$ receives no learning signal for actions not observed in the replay buffer, while this same Q-function is nevertheless queried on out-of-distribution actions during the policy update -- i.e., when $Q_\theta(s, a)$ is evaluated on $a\sim\pi(\cdot|s)$ in~\cref{eq:kl_control} -- and for bootstrapping critic targets in the squared TD-loss:
\vspace{-2mm}
\begin{equation}
\label{eq:tdloss}
\hspace{-2mm}J(Q_\theta) \defeq \hspace{-2mm}\E_{\substack{(s,a,s')\sim\D \\ a' \sim \pi(\cdot|s')}}[(r(s,a) + \gamma Q_{\hat{\theta}}(s',a')-Q_\theta(s,a))^2].\hspace{-1mm}
\end{equation}
Thus, issues with critic extrapolation can still dominate divergence regularizers applied to the policy.

\subsection{Critic Penalty}

Other works, such as AlgaeDICE ~\cite{nachum2019algaedice} and CQL~\cite{kumar2020conservative} attempt to incorporate some divergence regularization into the critic. %
In particular, AlgaeDICE introduces a term that pushes $Q$-values down for actions sampled from the training policy while minimizing TD-error via residual learning:
\vspace{-2mm}
\begin{align*}
\min_{\theta}   \alpha(1-\gamma)&\E_{\substack{s_0\sim\pi_0(\cdot)\\a_0\sim\pi(s_0)}}[Q_\theta(s_0,a_0)] +\\  \E_{(s,a)\sim\D}[(r(s,a) + \gamma &\E_{\substack{s'\sim p(\cdot|s,a)\\a'\sim\pi(\cdot|s')}}[Q_\theta(s',a')]-Q_\theta(s,a))^2]. 
\end{align*}
This formulation can be generalized by replacing the squared function by some convex function $f$. %
The choice of $f$ may be shown to correspond to an implicit $f$-divergence regularization on the learned policy with respect to state-action distributions.
For example, a choice of $f(x) = \exp(x - 1)$ or $f(x) = \log \E_{\D} \exp(x)$ corresponds to an implicit KL-divergence.

Based on a similar idea, CQL~\citep{kumar2020conservative} extends the standard critic loss $J(Q_\theta)$ in \cref{eq:tdloss} with additional terms that minimize $Q$-values sampled from a policy and maximize values of the dataset actions:
\begin{align}
\label{eqn:cql}
\hspace{-4mm}\min_{\theta} J(Q_\theta) + \lambda\hspace{-3mm}\E_{(s, a) \sim \D}\hspace{-1mm}[\log \Sigma_a \exp(Q_\theta(s,a)) - Q_\theta(s,a)].\hspace{-2mm}
\end{align}
Although both of these methods provide learning signal to the critic $Q$-values on the entire action space, AlgaeDICE is based on residual learning with slower convergence than fitted TD-learning~\citep{baird1995residual}. 
On the other hand, although CQL can achieve better empirical performance, the log-sum-exp term that appears in its formulation is not tractable for continuous actions and must be computed via numerical integration. %
The authors of CQL propose doing this via Monte-Carlo sampling with importance weights, where the current training policy is used to draw samples for the procedure. This process can add a significant computational burden to actor critic training. 
In contrast, our proposed Fisher-BRC will be shown to achieve favorable empirical performance while maintaining computational efficiency closer to standard actor critic.

\vspace{-2mm}
\section{Fisher-BRC}
\vspace{-2mm}
We now continue to describe our own approach to behavior regularization in offline RL settings, which aims to circumvent the issues associated with competing methods described in the previous section, namely (1) lack of well-defined critic values on out-of-distribution actions and (2) computational inefficiency of critic penalty approaches.
We begin with a conceptual derivation of our method, before presenting a more formal connection to Fisher divergence regularization.
See Algorithm~\ref{alg:pseudocode} for a sketch of the full algorithm.

\vspace{-2mm}
\begin{algorithm}[H]
\textbf{Input: } Dataset $\D$, offset network $O_\theta$, policy network $\pi_\phi$. 
\begin{enumerate}
    \itemsep0em 
    \item Learn approximate $\mu$ using behavioral cloning.
    \item Update $\theta$ using objective in~\cref{eq:bfrc}.
    \item Update $\phi$ using entropy-regularized objective \\ $\E_{s\sim\D,a\sim\pi_\phi(\cdot|s)}[O_\theta(s,a)+\log\mu(a|s) + \alpha \mathcal{H}(\pi_\phi(\cdot|s))]$.
    \item Repeat from 2. 
\end{enumerate}
\textbf{Return:} $\pi_\phi$.
\caption{Fisher-BRC [Sketch].}
\label{alg:pseudocode}
\end{algorithm}

\subsection{Conceptual Derivation}
\label{sec:concept}
We start from the observation that we can represent the entropy smoothed $Q$-values of the behavior policy $\mu(\cdot|s)$ as 
\begin{equation*}
Q(s,a) = V(s) + \log \mu(a|s).
\end{equation*}
This decomposition of $Q$-values into state-value and log-policy is popular in the entropy-regularized online RL literature~\citep{peters2010relative,nachum2017bridging}, where the $\mu$ in our notation is treated as the policy $\pi$ to be learned.
Our own setting is markedly different from these previous approaches in that  $\mu$ is the behavior policy and is fixed. Nevertheless, what would happen if we were to parameterize $Q$-values in this way, i.e, as
\begin{equation}
\label{eq:q-soft}
Q_\theta(s,a) := V_\theta(s) + \log \mu(a|s),
\end{equation}
and then learn via standard TD error minimization~\eqref{eq:tdloss}? Well, there is an advantage, but also a disadvantage.

First, the main advantage of this formulation is that, in contrast to a $Q$-function parameterized by its own neural network function approximator, the density $\mu(a|s)$ is well-defined for all actions, and thus $Q$ is more likely to have a well-behaved landscape even though its training may only cover a small subset of the full action space. Accordingly, the learned policy $\pi$, when trained in the standard way to choose actions which maximize the $Q$-values, is thus encouraged to stay close to $\mu$, without the need for an explicit divergence penalty as in~\eqref{eq:kl_control}.
In practice, knowledge of $\mu$ is not explicitly provided, and one usually resorts to behavioral cloning on the offline dataset to approximate $\mu$~\citep{wu2019behavior}. 
Nevertheless, the advantage of the formulation in~\eqref{eq:q-soft} still holds, since even if $\mu$ is trained on a sparse subset of all possible actions, the fact that it is a normalized probability distribution means that $\mu(a|s)$ assigns low probabilities to actions outside of $\D$.
Still, in practice some density models might fail to generalize to out-of-distribution data \cite{kirichenko2020normalizing}. For this reason, we parameterize the approximate density $\mu$ as a mixture density model \cite{bishop1994mixture}. %

As for the disadvantage, it is clear that the formulation of $Q$ in~\eqref{eq:q-soft} is too restrictive. If this representation is used for training a new policy $\pi$, the new policy will be limited to copying the behavior policy $\mu$, regardless of $V_\theta$, and this will lead to suboptimal performance. %
In order to address this issue and enable the learned $\pi$ to generalize beyond just mimicking $\mu$, we propose replacing the value function $V_\theta(s)$ with a state-action value \emph{offset} function $O_\theta(s,a)$: %
\begin{equation}
\label{eq:qrepr}
    Q_\theta(s,a) \defeq O_\theta(s,a) + \log \mu(a|s).
\end{equation}
With this representation one can learn a richer representation of $Q$-values. %
However, this parameterization can potentially put us back in the fully-parameterized $Q_\theta$ regime of vanilla actor critic. It is clear that the offset term must be suitably constrained or regularized to find an appropriate middle-ground between the overly-restrictive $V_\theta(s)$ and the fully-parameterized alternative that is liable to extrapolation errors and policy divergence.

To motivate what an appropriate regularization on $O_\theta$ should be, we begin by dissecting exactly how the offset term impacts the learned policy. Let's take a closer look at the policy updates in standard continuous-control actor critic~\citep{lillicrap2019continuous,haarnoja2019soft}. These updates are based on the chain-rule gradient computation below:
\begin{multline}
    \label{eq:inc_update}
    \nabla_\phi Q_\theta(s, \pi_\phi(s)) = \\
    \big[\nabla_a O_\theta(s,a) + \nabla_a \log \mu(a|s)\big]_{a = \pi_\phi(s)} \nabla_\phi \pi_\phi(s).
\end{multline}
From this expression, it is clear that potential extrapolation issues with the actor arise via the gradient $\nabla_a O_\theta(s,a)$. That is, without appropriate constraints or regularizers on $O_\theta(s,a)$, its gradients might dominate over the gradients of the behavior policy $\nabla_a \log \mu(a|s)$. Therefore, as a way to control the trade-off between over-constraining $\pi$ to be close to the behavior policy and learning more rich representations of $Q$-values, we propose using a gradient penalty regularizer of the form $\|\nabla_a O_\theta(s,a)\|^2$. 
Accordingly, the full critic optimization objective is as follows:
\begin{align}
\label{eq:bfrc}
\min_{\theta} J(O_\theta + \log\mu) + \lambda \E_{\substack{s \sim \D \\a\sim\pi_\phi(\cdot|s)}}[\|\nabla_a O_\theta(s,a) \|^2], 
\end{align}
where $J(\cdot)$ is defined as in \cref{eq:tdloss} and $\mu$ is not updated during critic learning. In this objective, $\lambda$ is a hyperparameter that controls the contribution of the gradient penalty term.
Unless otherwise noted, we set $\lambda=0.1$ as the regularization coefficient.

A keen reader may note that the gradients in \cref{eq:inc_update} look similar to gradients of \cref{eq:kl_control}, with the difference that we take gradients of the offset function instead of the critic function.
However, in our setting the critic loss is substantially different, since the offset term plus the log behavior density learns to predict the \emph{unmodified} $Q$-values of the training policy instead of $Q$-values augmented with a KL-divergence term. 
For example, if the rewards of the MDP already naturally compel the learned policy to match the behavior policy, the offset term in Fisher-BRC can vanish, whereas the explicit divergence penalty in BRAC will bias the learned $Q$-values unnecessarily.

\label{sec:fbrc}

\subsection{Fisher Divergence Derivation}
We now show how the same objective in~\eqref{eq:bfrc} may be derived from the perspective of Fisher divergence regularization.
We begin by introducing the idea of Boltzmann policies -- essentially policies expressed as a Boltzmann distribution with energy function given by a set of $Q$-values. 
We then present the Fisher divergence, and show how a Fisher divergence regularizer between a Boltzmann policy and the behavior policy reduces to the gradient penalty proposed in~\eqref{eq:bfrc}.
Finally, we elaborate on connections to CQL, which we show can be interpreted as a KL divergence regularization between the Boltzmann policy and the behavior policy, and this insight may be of independent interest, since the original derivation of CQL is via a very different route.

\vspace{-3mm}
\paragraph{Boltzmann Policies} %

For a given $Q$-value function, the associated Boltzmann policy is given by the following expression:
\vspace{-4mm}
\begin{equation}
\label{eqn:bolt}
    \pi_{ebm}(a|s) \defeq \cfrac{\exp(Q(s,a))}{\sum \exp(Q(s,a))}.
\end{equation}
In an actor critic setting, the actor will recover this same policy if an entropy regularizer is added to the actor loss, as is commonly done~\citep{haarnoja2019soft}.
While there are some works which use this representation of a policy more explicitly~\citep{fox2015taming,haarnoja2017reinforcement,nachum2017bridging}, a main disadvantage is that the normalization term may not be tractable for continuous actions, and so computing the policy distribution $\pi_{ebm}$ explicitly %
requires performing computationally expensive numerical integration.
Thus, it is more common to only recover the policy via entropy regularization on the actor loss.

\vspace{-3mm}
\paragraph{Fisher Divergence} In order to avoid issues with computing the normalization term, we consider the Fisher divergence, or Fisher information distance \cite{johnson2004information}:
\begin{equation}
\label{eqn:fisher}    
\hspace{-2mm}F(p(\cdot), q(\cdot)) = \hspace{-2mm}\E_{x \sim p(\cdot)} \left[ \|\nabla_x \log p(x) - \nabla_x \log q(x) \|^2 \right].\hspace{-2mm}
\end{equation}
As one can see from the formulation, in order to compute the Fisher divergence between two distributions, we need only have sampling access to $p(x)$ and the ability to compute $\nabla_x \log p(x),\nabla_x \log q(x)$. Crucially, the computation of either $\nabla_x \log p(x)$ or $\nabla_x \log q(x)$ does not require normalized distributions, since the normalization term (which is a constant with respect to $x$) disappears from $\log p(x),\log q(x)$ due to the differentiation. Thus, we can avoid computing the normalization term in \eqref{eqn:bolt}.

Since the Fisher divergence is amenable to Boltzmann representations of policies, let's consider an optimization objective that consists of a squared TD-loss and a Fisher divergence term between the Boltzmann distribution and the behavior policy $\mu$:
\begin{align}
\begin{split}
\label{eqn:fisher_td}
&J(Q_\theta) + \lambda \E_{s\sim \D} \left[F\left( \cfrac{\exp(Q(s,\cdot))}{\sum_a \exp(Q(s,a))},\mu(\cdot|s)\right)\right] =\\ 
&J(Q_\theta) + \lambda\E_{\substack{s \sim \D \\ a \sim \pi_{emb}(\cdot|s)}} \left[\left\|\nabla_a \log \mu(a|s)- \nabla_a Q(s,a) \right\|^2 \right].
\end{split}
\end{align}
\vspace{-2pt}
The coefficient $\lambda$ controls the strength of the Fisher divergence term. 
We can further simplify this objective by using the representation of $Q$ proposed in \eqref{eq:qrepr}:
\vspace{-2mm}
\begin{align}
&J(O_\theta + \log\mu) + \lambda\E_{\substack{s \sim \D \\ a \sim \pi_{ebm}(\cdot|s)}} \left[\left\|\nabla_a O_\theta(s,a) \right\|^2 \right].
\label{eq:fisher-ebm}
\end{align}
The only difference with \cref{eq:bfrc} is that actions are sampled from the training policy $\pi_\phi$, while in this case the actions are sampled from the Bolzman policy $\pi_{ebm}$. In practice, sampling from $\pi_{ebm}$ can be just as computationally expensive as computing the normalization term in \cref{eqn:cql}. 
However, as mentioned earlier, the Boltzmann policy $\pi_{ebm}$ may be recovered by the actor via incorporation of an entropy regularizer in the actor loss. 
Thus, we propose to train our actor $\pi_\phi$ exactly in this manner (which is already popular in model-free actor critic algorithms~\citep{haarnoja2019soft}), and then use it in~\cref{eq:fisher-ebm} as a plug-in approximation of $\pi_{ebm}$.
In this way, we have arrived again to the same objective first defined in \cref{sec:fbrc}.

\begin{figure*}[ht!]
    \centering
    \includegraphics[width=0.33\textwidth]{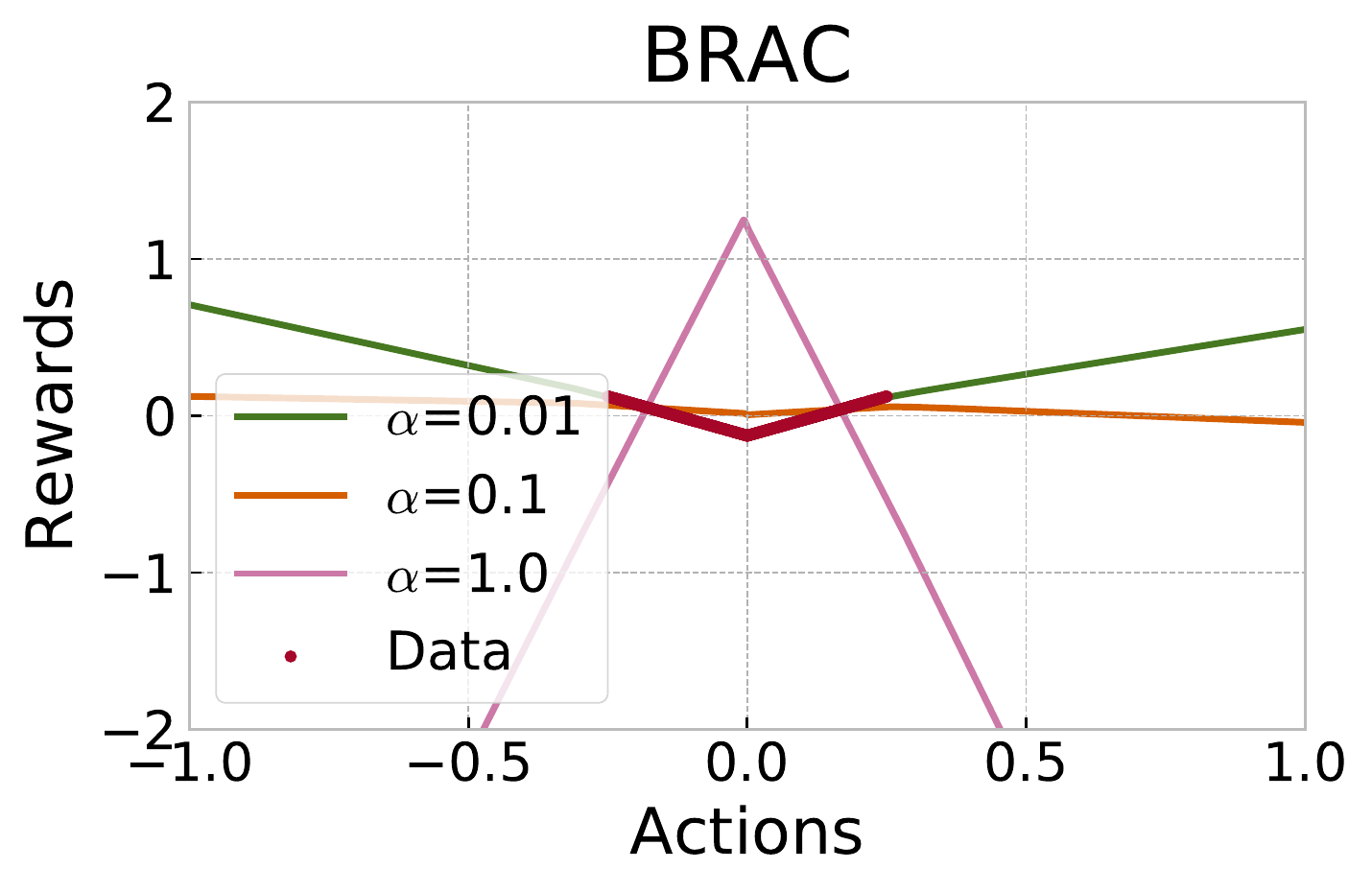} 
    \includegraphics[width=0.33\textwidth]{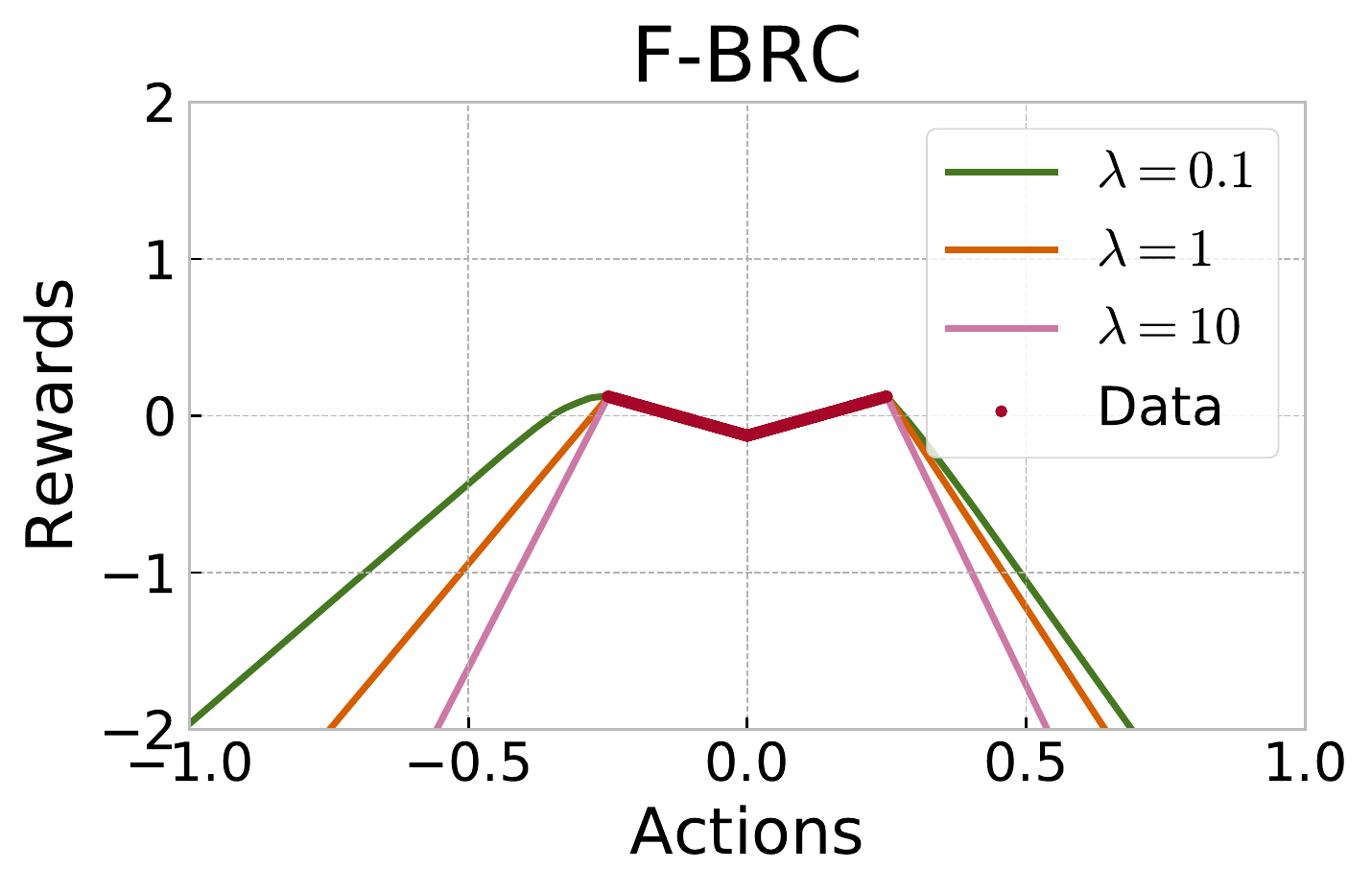}
\caption{The objective landscapes (the regularized critic values) for the policy induced by the learned critic in BRAC or the parameterized offset critic in Fisher-BRC.
The observed actions in the offline data are all within $[-0.25, 0.25]$, and suggest the optimal reward-maximizing actions as $\{-0.25, 0.25\}$. 
In BRAC (left), we see the landscape is heavily dependent on the choice of KL-divergence coefficient $\alpha$, and it is easy to either over-regularize (with optimum around $0.0$) or over-extrapolate (with optima far from the observed actions in $[-0.25, 0.25]$). On the other hand, due to the unique parameterization used in Fisher-BRC critic, its corresponding objective landscape correctly predicts the pessimistic reward values and peaks at the modes of the true reward function (right). We also see that Fisher-BRC is more robust to choice of regularizer coefficient $\lambda$.}    \label{fig:toy}
\end{figure*}

\begin{table*}[t]
\setlength{\tabcolsep}{3pt}
\centering
\begin{tabular}{lrrrrrr||r}
\toprule
{} &    BC &  BRAC-p &  BRAC-v &  MBOP &   
CQL (GitHub) & CQL (Ours) 
&  F-BRC (Ours) \\
\midrule
halfcheetah-random        &  $30.5$ &    $23.5$ &    $28.1$ &   $6.3 \pm 4.0$ & $27.1 \pm 1.3$ & $20.7 \pm 0.6$ &  $33.3 \pm 1.3$ \\
hopper-random             &  $11.3$ &    $11.1$ &    $12.0$ &  $10.8 \pm 0.3$ &  $10.6 \pm 0.1$ & $10.4 \pm 0.1$ & $11.3 \pm 0.2$ \\
walker2d-random           &   $4.1$ &     $0.8$ &     $0.5$ &   $8.1 \pm 5.5$ &   $1.1 \pm 2.2$ & $10.0 \pm 4.6$ &  $1.5 \pm 0.7$
 \\ \hline
halfcheetah-medium        &  $36.1$ &    $44.0$ &    $45.5$ &  $44.6 \pm 0.8$ &  $40.3 \pm 0.3$ & $38.9 \pm 0.3$ &  $41.3 \pm 0.3$ \\
walker2d-medium           &   $6.6$ &    $72.7$ &    $81.3$ &  $41.0\pm29.4$ & $77.3 \pm 3.8$ & $69.2 \pm 8.3$ & $78.8 \pm 1.0$ \\
hopper-medium             &  $29.0$ &    $31.2$ &    $32.3$ &  $48.8\pm26.8$ &  $42.2 \pm 15.5$ &$30.5 \pm 0.7$ &  $99.4 \pm 0.3$ \\ \hline
halfcheetah-expert        & $107.0$ &     $3.8$ &    $-1.1$ &   - & $54.4 \pm 45.8$ & $103.5 \pm 1.3$ & $108.4 \pm 0.5$ \\
hopper-expert             & $109.0$ &     $6.6$ &     $3.7$ &   - & $67.7 \pm 54.7$ &  $112.2 \pm 0.2$ & $112.3 \pm 0.1$ \\
walker2d-expert           & $125.7$ &    $-0.2$ &    $-0.0$ &   - & $84.7 \pm 42.7$ & $107.2 \pm 3.8$& $103.0 \pm 5.0$ \\ \hline
halfcheetah-medium-expert &  $35.8$ &    $43.8$ &    $45.3$ & $105.9\pm17.8$ & $21.7 \pm 6.8$ & $58.6 \pm 8.7$&   $93.3 \pm 10.2$ \\
walker2d-medium-expert    &  $11.3$ &    $-0.3$ &     $0.9$ &  $70.2\pm36.2$ & $104.0 \pm 10.1$ & $104.6 \pm 10.4$ & $105.2 \pm 3.9$ \\
hopper-medium-expert      & $111.9$ &     $1.1$ &     $0.8$ &  $55.1\pm44.3$ & $111.3 \pm 2.1$ & $112.4 \pm 0.2$& $112.4 \pm 0.3$ \\ \hline
halfcheetah-mixed         &  $38.4$ &    $45.6$ &    $45.9$ &  $42.3\pm0.9$ &  $44.9 \pm 1.1$ & $42.0 \pm 1.1$ &   $43.2 \pm 1.5$ \\
hopper-mixed              &  $11.8$ &     $0.7$ &     $0.8$ &  $12.4\pm5.8$ &  $31.6 \pm 3.6$ & $29.0 \pm 0.5$ & $35.6 \pm 1.0$ \\
walker2d-mixed            &  $11.3$ &    $-0.3$ &     $0.9$ &   $9.7\pm5.3$ &  $16.8 \pm 3.1$ & $16.5 \pm 4.9$ & $41.8 \pm 7.9$ \\
\bottomrule
\end{tabular}
\caption{Comparison of our method (F-BRC) to prior work. The results for BC and BRAC are taken from~\citet{fu2020d4rl}; the results for MBOP are taken from~\citet{argenson2020model}; the results for CQL (GitHub) are taken from the author-provided open-source implementation of ~\citep{kumar2020conservative}; and the results for CQL (Ours) are from our own re-implementation of CQL. For all methods we run ourselves, we plot the normalized returns at the end of training (without early stopping) computed over 5 seeds. For every seed we run evaluation for 10 episodes.
}
\label{tab:results}
\end{table*}

The connection of our proposed objective to the Fisher divergence recalls similar quantities in the score matching and energy-based generative model literature.
In fact, the Fisher divergence is a popular metric in these literatures exactly because it avoids an expensive computation of a log-normalizer, which is necessary for other common divergences~\citep{lyu2012interpretation,bao2020variational}.
Moreover, many works in the generative model literature employ gradient penalties, even if they do not explicitly make a connection to the Fisher divergence.
This provides a further empirical advantage to our method, as these gradient penalties -- due to their popularity -- may be efficiently implemented using many modern machine learning libraries. 
We note that although we use a soft penalty, one can enforce hard constraints as in Spectral Normalized GANs \cite{miyato2018spectral}, but we leave this for future work.

Due the connection of our method to Fisher Divergence, we dub our method Fisher-BRC (Fisher Behavior Regularized Critic).

\paragraph{Connections to CQL} The CQL objective, although originally derived in~\citet{kumar2020conservative} from a very different perspective, can also be motivated as a regularizer on a Boltzmann policy. In the case of CQL, the regularizer is the common KL-divergence:
\begin{equation}
\label{eqn:cql_kl}
\hspace{-10pt} J(Q_\theta) + \lambda\E_{s\sim\D} \left[D_{KL}\left(\mu(\cdot|s)\middle|\cfrac{\exp(Q(s,\cdot))}{\sum_a \exp(Q(s,a))}\right)\right]. 
\end{equation}
Expanding the KL-Divergence term yields, 
\begin{align*}
    D_{KL}(\mu(\cdot|s)|\pi_{emb}(\cdot|s)) = \E_{a\sim\mu(\cdot|s)}\left[\log \cfrac{\mu(a|s)}{\pi_{emb}(a|s)}\right]= \\
    \E_{a\sim\mu(\cdot|s)}\left[\log \sum_a \exp(Q(s,a)) - Q(s,a) + \log \mu(a|s)\right],
\end{align*}
and this is equivalent to the familiar form of CQL from~\eqref{eqn:cql}, since $\log \mu(a|s)$ is a constant with respect to $Q$.

As mentioned earlier, for continuous distributions the normalization term in CQL is not tractable and necessitates expensive numerical integration. In CQL this integration is calculated by sampling from the training policy and importance weighting. Thus, our own method enjoys a significant computational advantage. Our use of a novel critic representation and the Fisher divergence allows us to circumvent this practical issue.

We note that in this derivation of CQL, the direction of divergence in~\cref{eqn:cql_kl} is switched compared to \cref{eqn:fisher_td} in Fisher-BRC. 
One may derive a variant of Fisher-BRC using this same direction of the divergence, and this will result in the same expression in \cref{eq:bfrc}, only that the expectation of the gradient penalty is switched to be over $(s,a) \sim \D$.
Anecdotally, in initial experiments we did not observe large empirical differences when training with this alternative objective. However, due to the closer connection to the actor loss when using $a\sim \pi_\phi(\cdot|s)$ (see \cref{sec:concept}), we stick to the formulation originally presented in \cref{eq:bfrc}.

\section{Experiments}
We present empirical demonstrations of Fisher-BRC in a variety of settings.
We start with a simple continuous bandit experiment that illustrates the difference between our method and more common behavior regularization techniques based on explicit divergence penalties. %
Then we evaluate our method against state-of-the-art offline RL model-based and model-free algorithms on the D4RL benchmark datasets.
Finally, we analyze the effect of gradient penalty regularization and provide statistics on the computational advantage of Fisher-BRC compared to CQL.

\subsection{Toy Continuous Bandit Problem}
We begin with a simple conceptual demonstration comparing Fisher-BRC to the similar and common alternative of explicit divergence penalties applied to the learned policy.
Specifically, we consider the loss in~\cref{eq:kl_control}, which corresponds to the policy update used in BRAC~\citep{wu2019behavior}.

For the purpose of this experiment, we consider a continuous bandit
with one-dimensional action space given by $[-1,1]$. The rewards are given by 
\[
    r(a)= 
\begin{cases}
    |a| - 0.125,& \text{if } a \in [-0.25, 0.25] \\
    -\infty,              & \text{otherwise.}
\end{cases}
\]
The offline training dataset is collected by sampling 1000 actions from a uniform distribution, $a \sim \U(-0.25, 0.25)$, and recording the corresponding rewards $r(a)$. 
Thus, the distribution of actions exhibits poor coverage of the full action space, and the rewards for $a \in [-1, -0.25] \cup [0.25, 1]$ are not observed in the dataset.

Both BRAC and Fisher-BRC require a fitted behavior policy. To do so, we fit a behavior policy $\mu(a)$ parameterized as a Laplace distribution.\footnote{Our choice of a Laplace parameterization is to match the absolute value appearing in the definition of rewards $r(a)$. If a Gaussian parameterization is used, then the rewards may be modified to a quadratic function to achieve the same result.} 
Subsequently, we fit the critic for BRAC and F-BRC.
In BRAC, we fit a critic using mean squared error to match the rewards: $\E_{(a, r)\sim\D}[(R_\theta(a) - r)^2]$. 
For Fisher-BRC, we use the representation and regularization from \cref{eq:bfrc}, assuming an initialization of $\pi$ to $\U$: %
\begin{align*}
\E_{(a, r)\sim\D}[(O_\theta(a) + \log\mu(a) - r)^2] + \\ \lambda \E_{a_{reg}\sim \U(-1, 1)} \|\nabla_a O_\theta(a_{reg})\|^2.
\end{align*}
These critics then determine the objective landscape for the learned policy. 
We plots these landscapes in~\cref{fig:toy}. 
Specifically, we plot $R_\theta(a) + \alpha \log\mu(a)$ for BRAC and $O_\theta(a) + \log \mu(a)$ for Fisher-BRC for a variety of choices of $\alpha$ and $\lambda$. 
Recall that the policy will be learned to choose actions which maximize these values. 
Thus, ideally these objective landscapes should possess optima around the globally optimal actions $\{-0.25, 0.25\}$.

We observe that it is hard to pick the KL-coefficient $\alpha$ for the policy loss landscape induced by BRAC to avoid either over generalizing (with optima outside of $[-0.25,0.25]$) or over regularizing (with optima far from the optimal actions $\{-0.25, 0.25\}$); see \cref{fig:toy}, left. 

\begin{figure}[ht!]
    \centering
    \includegraphics[width=0.41\textwidth]{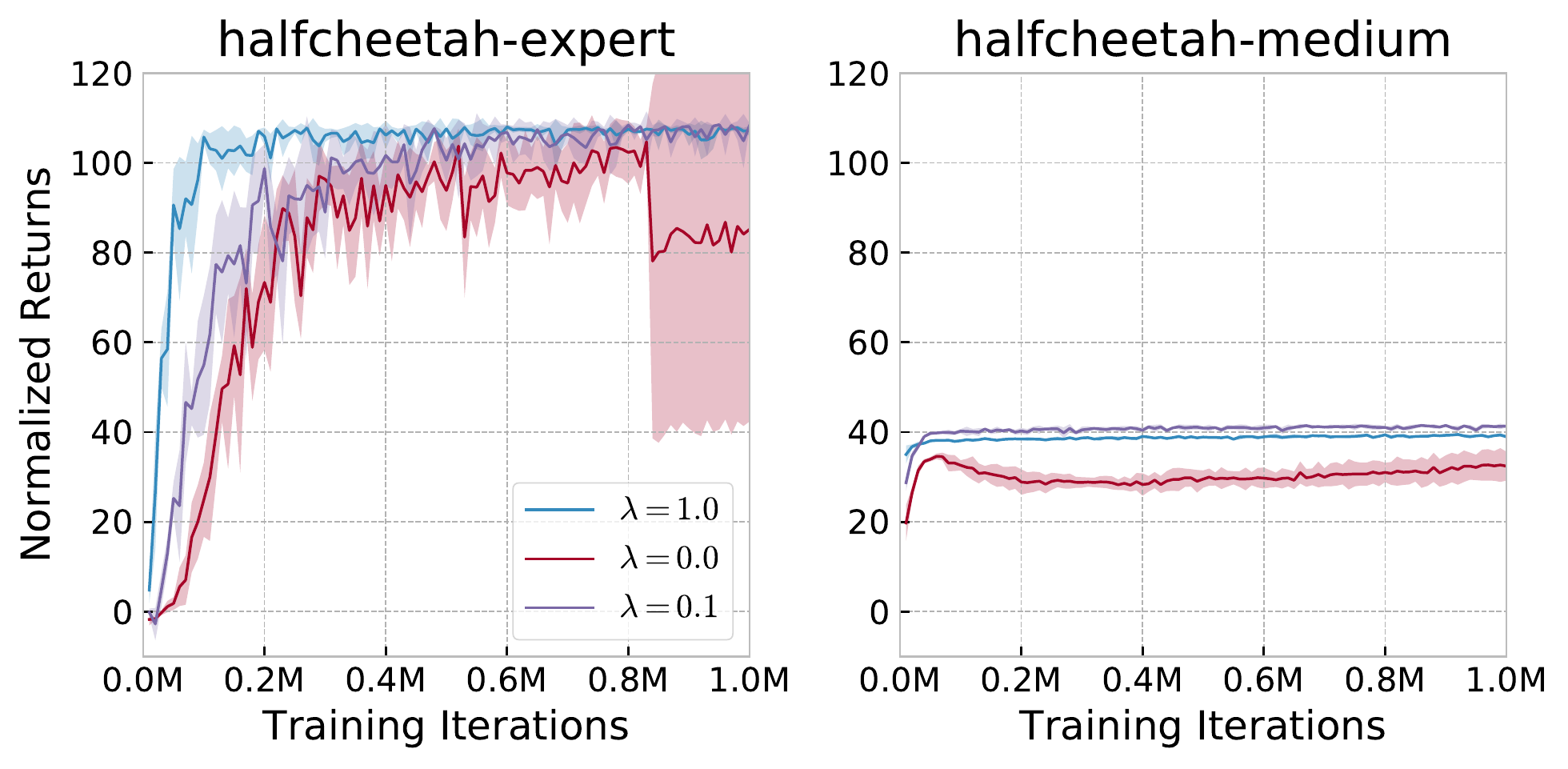}
    \includegraphics[width=0.41\textwidth]{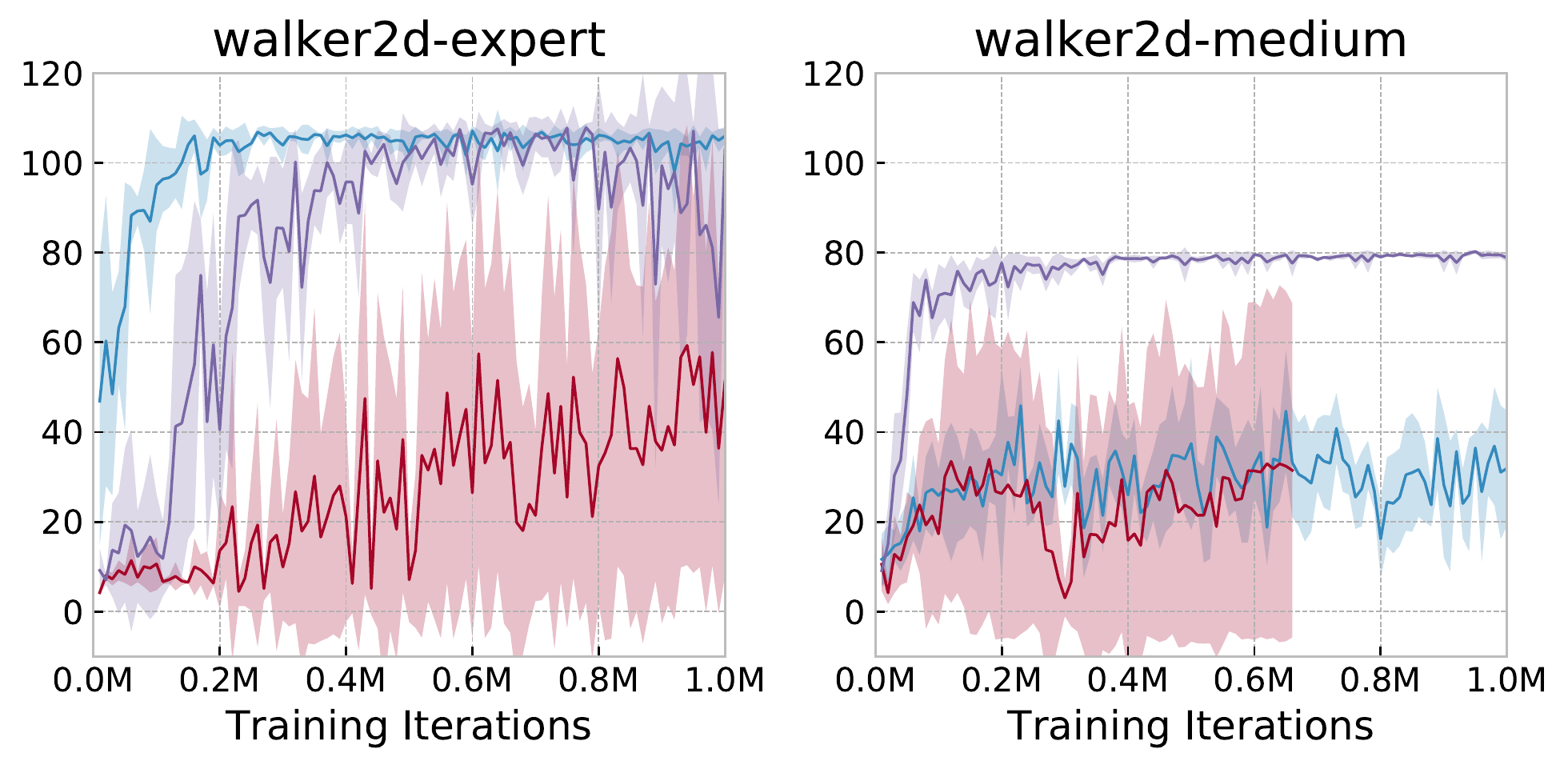}
    \caption{Performance of F-BRC for different values of the gradient penalty coefficient. A larger value, $\lambda=1$, over-constraints the learned policy to stay close to the behavior policy. This leads to more stable performance on expert datasets, where the behavior policy is near-optimal, but worse performance on medium datasets. Without the regularization ($\lambda=0.0$) Fisher-BRC collapses on most of these tasks; when the plot is cutoff, it means at least one of the seeds produced NaN values in training.}
    \label{fig:abl_plots}
\end{figure}

\begin{figure*}%
    \centering
    \includegraphics[width=\textwidth]{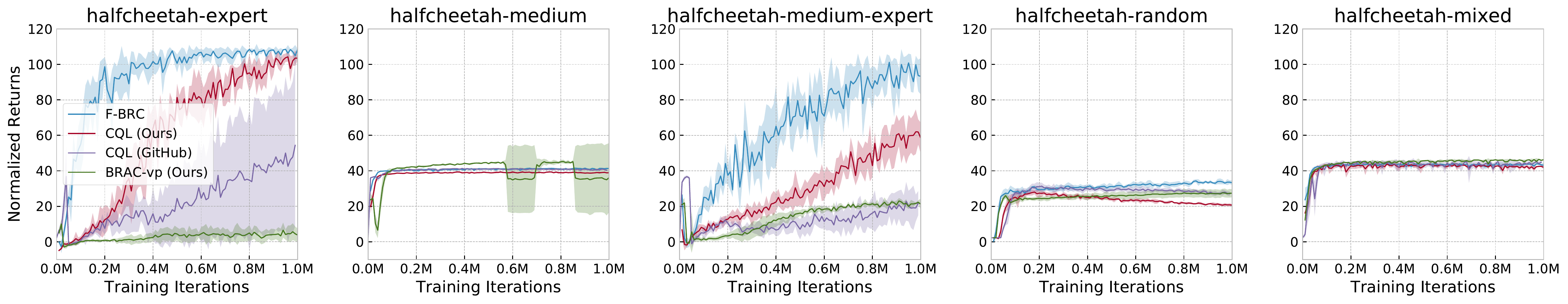} \\
    \includegraphics[width=\textwidth]{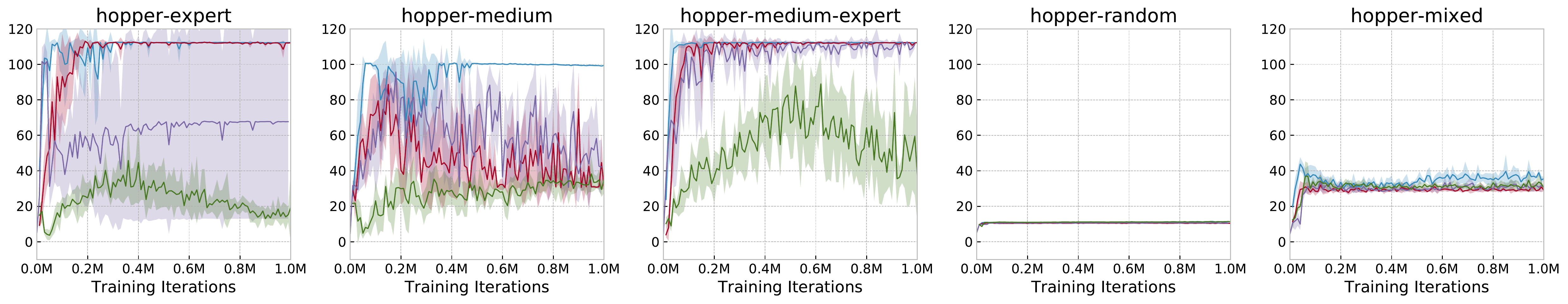} \\
    \includegraphics[width=\textwidth]{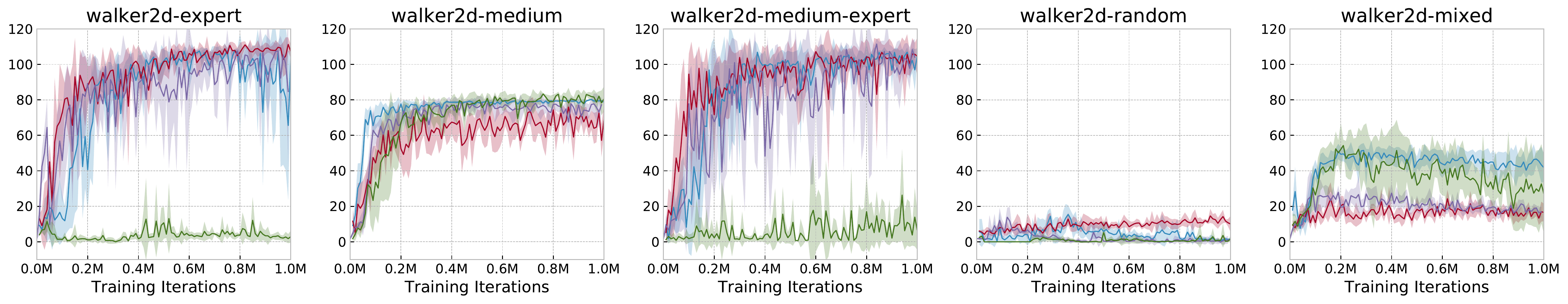}
    \caption{We compare F-BRC against prior methods in terms of convergence speed with respect to gradient updates steps. We see that Fisher-BRC enjoys better final performance and faster convergence in most tasks compared to BRAC and CQL.}
    \label{fig:plots}
\end{figure*}

On the other hand, Fisher-BRC correctly constrains the learned value function -- in fact, the value function is unmodified for in-distribution samples -- and is robust to the choice of the regularization hyperparameter (\cref{fig:toy}, right).

\subsection{Deep Offline RL Benchmarks}

We continue to present Fisher-BRC on more complex environments.
We compare our method to prior work on the OpenAI Gym MuJoCo tasks using D4RL datasets~\citep{fu2020d4rl}. 
We consider the following baselines: BRAC-vp and BRAC-pr~\citep{wu2019behavior}, due to a similar policy learning objective, MBOP \cite{argenson2020model}, the state-of-state in offline model-based reinforcement learning, and CQL~\citep{kumar2020conservative}, due to a similar connection with energy based learning. 
Our implementation for Fisher-BRC follows the standard SAC implementation, only that we use a 3-layer network as in CQL. Additional implementation details are in the appendix.

\vspace{-3mm}
\paragraph{Overall performance}
The results of our method and all considered baselines are presented in Table~\ref{tab:results}.
Our method performs comparably or surpasses prior work on most of the tasks. Notably, many of the baseline algorithms exhibit inconsistent performance across the tasks, achieving good performance on some while poor performance on others. In contrast, our Fisher-BRC exhibits consistent and good performance across almost all tasks.

\vspace{-3mm}
\paragraph{Effect of gradient penalty} As a way of investigating the effect of gradient penalty vs. the offset parameterization in Fisher-BRC, we evaluate different values of gradient penalty. 
We present results of $\lambda \in \{0.0, 0.1, 1.0\}$ in~\cref{fig:abl_plots}.
We note that the gradient penalty is an important component of Fisher-BRC, since when $\lambda=0.0$ performance degrades dramatically.
On the other hand, when $\lambda$ is set too high ($\lambda=1.0$), we see that the learned policy is over-constrained; i.e., we see that performance on the expert datasets is improved while performance is limited on the medium datasets, since the behavior policy in these datasets is highly sub-optimal.
This is expected, since a high $\lambda$ corresponds to a large gradient penalty on the regularization on the offset, compelling the offset to be near-constant with respect to actions.

\vspace{-3mm}
\paragraph{Convergence speed} %
We also evaluate the wall-clock training time of our method compared with CQL, which demonstrates comparable policy performance on the D4RL tasks but significantly different computational efficiency. The total training time for 1 million steps for Fisher-BRC is 1.4 hours of behavioral cloning pretraining followed by 6.2 hours of policy training. CQL total training time is 16.3 hours (which does not require pre-training of a behavior density policy).
Our method converges faster not only in terms of gradient updates (see \cref{fig:plots}), but it is also computationally faster due to omitting the expensive numerical integration to compute the log-sum-exp term of CQL. These experiments were carried out on a Google cloud instance containing an AMD EPYC 7B12 CPU at 2.25GHz (using 8 of 64 available cores) and 32GB of RAM.

\section{Conclusions}
We have introduced Fisher-BRC, a simple critic representation and regularization technique for offline reinforcement learning. 
Our derivations highlight connections between our training objective and Fisher divergence regularization from score matching and energy-based model literature.
Our method is easy to implement and highly performant.
Compared to existing offline RL algorithms, Fisher-BRC exhibits better and more consistent performance across a variety of domains.

\section*{Acknowledgements} We thank George Tucker for reviewing a draft of this paper and providing valuable feedback. We also thank Aviral Kumar for discussions and help with CQL results.

\bibliography{fbrac}
\bibliographystyle{icml2021}

\clearpage

\appendix
\onecolumn
\section{Implementation Details}

\paragraph{Behavior policy} We fit the behavior model using a conditional Mixture of Gaussians~\cite{bishop1994mixture} with tanh squashing~\cite{haarnoja2017reinforcement}. We use 5 mixture components.
We train the density model with Adam optimizer \cite{kingma2014adam} for $10^6$ steps and starting from learning rate $10^{-3}$ and decreasing it by 10 at $8\cdot10^5$ and $9\cdot10^5$ gradient update steps. Similar to BRAC we train the behavior actor with SAC-style entropy regularization with the same target entropy. We parameterize the model as a 3 layer MLP with relu activations and 256 hidden units. 

\paragraph{Actor and critic learning} Our implementation is based on Soft Actor Critic \cite{haarnoja2019soft}. As in CQL we do not add entropy to the rewards and we modify the critic loss to accommodate the additional regularization term. We use default SAC hyper parameters without additional tuning, in contrast to CQL and BRAC which tune policy learning rate. Following CQL we increased network size for the actor and the critic to 3 layer MLP with 256 hidden units.

\paragraph{Survival bonus} The linear term used in CQL can be seen as adding a survival bonus for the environments with early early termination.  The derivation is included in the Appendix. Adding a positive constant to the rewards does not have an effect on the optimal policy in infinite horizon MDPs, but in practice Q-targets for terminal states are replaced with $0$ that leads to having either a survival bonus or step penalty. For this reason, we add a reward bonus to our implementation as well for fair comparison. We choose the same value $\lambda_{cql}=5$ as in CQL.

In particular, one can verify that 
\begin{align*}
     \nabla_\theta [-\lambda_{cql} &Q_\theta(s,a) + (\gamma Q_{\hat{\theta}}(s',a') + r(s,a) - Q_\theta(s,a))^2]= \\
    -\lambda_{cql}\nabla_\theta &Q_\theta(s,a) -
    (\gamma Q_{\hat{\theta}}(s',a') + r(s,a) - Q_\theta(s,a)) \nabla_\theta Q_\theta(s,a)=\\
    - (\gamma Q_{\hat{\theta}}(s',&a') + [r(s,a) + \lambda_{cql}] - Q_\theta(s,a)) \nabla_\theta Q_\theta(s,a)=\\
     \nabla_\theta (\gamma &Q_{\hat{\theta}}(s',a') + [r(s,a) + \lambda_{cql}] - Q_\theta(s,a))^2.
\end{align*}

\newpage

\section{Gradient Penalty Ablation}

We also present a full set of results for the ablations in \cref{fig:abl_plots}.

\begin{figure*}[h]
    \centering
    \includegraphics[width=\textwidth]{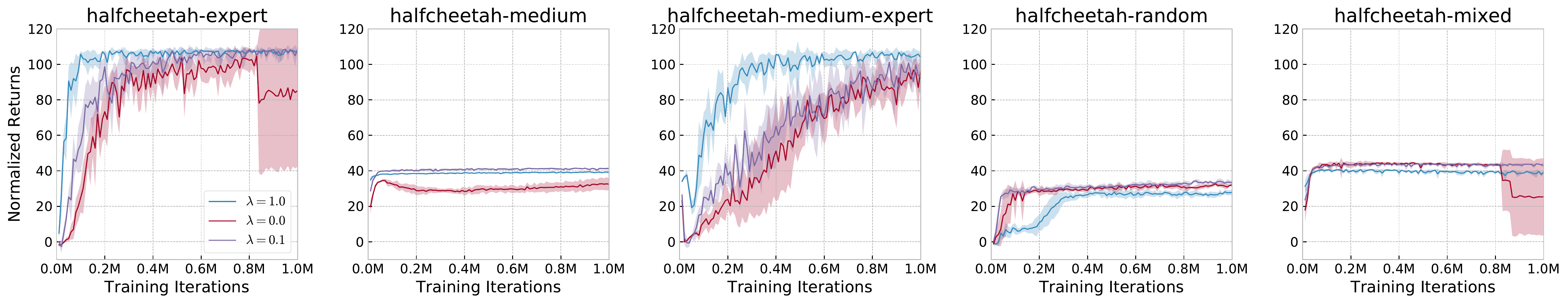} \\
    \includegraphics[width=\textwidth]{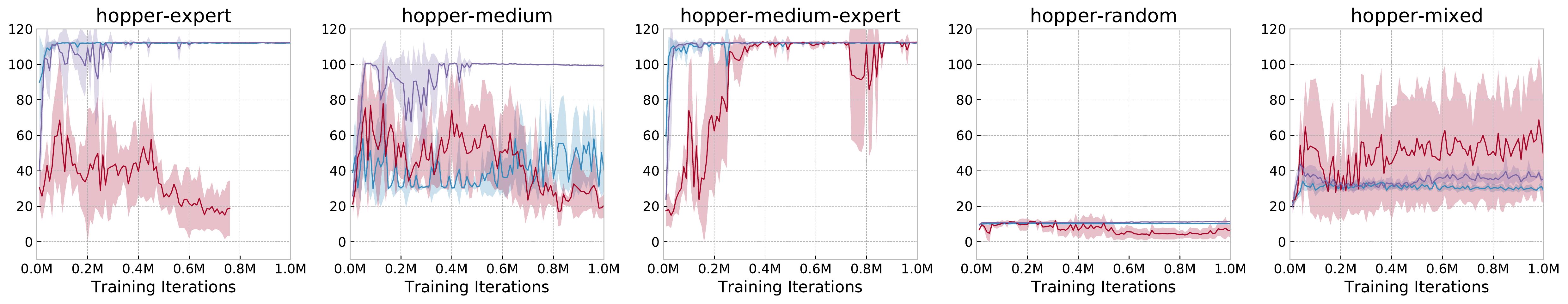} \\
    \includegraphics[width=\textwidth]{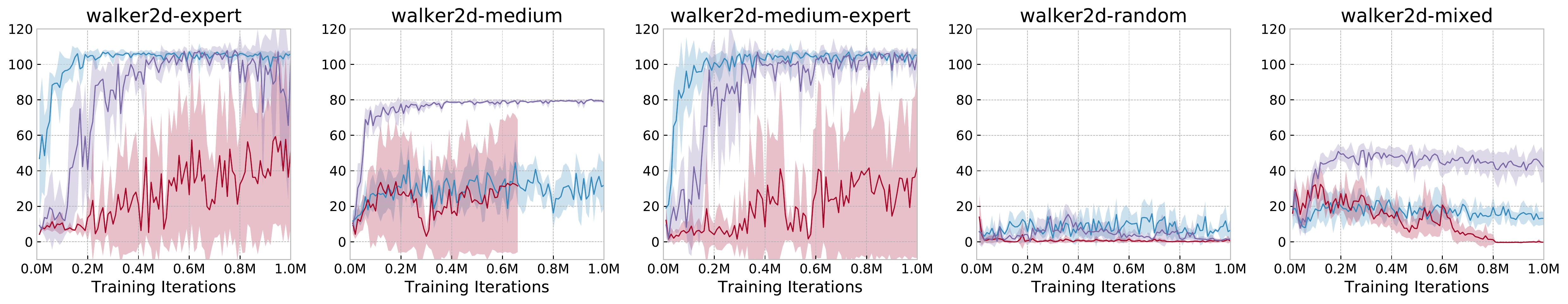}
    \caption{Performance of F-BRC for different values of the gradient penalty coefficient. A larger value, $\lambda=1$, over-constraints the learned policy to stay close to the behavior policy. This leads to more stable performance on expert datasets, where the behavior policy is near-optimal, but worse performance on medium datasets. Without the regularization ($\lambda=0.0$) Fisher-BRC collapses on most of these tasks; when the plot is cutoff, it means at least one of the seeds produced NaN values in training.}
    \label{fig:abl_plots_}
\end{figure*}

\newpage

\section{Critic Regularization Ablation}

We also evaluate the effect of gradient penalty of standard Soft Actor Critic without the critic representation introduced in this paper.

\begin{figure*}[h]
    \centering
    \includegraphics[width=\textwidth]{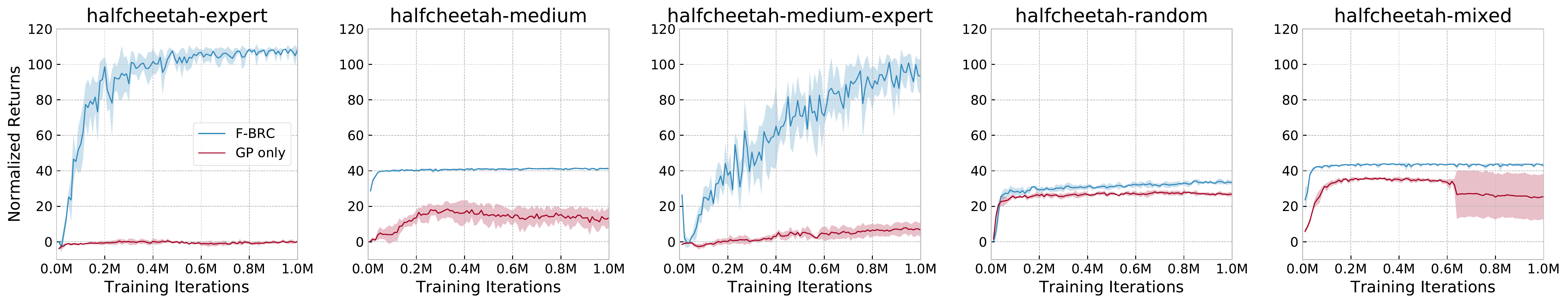} \\
    \includegraphics[width=\textwidth]{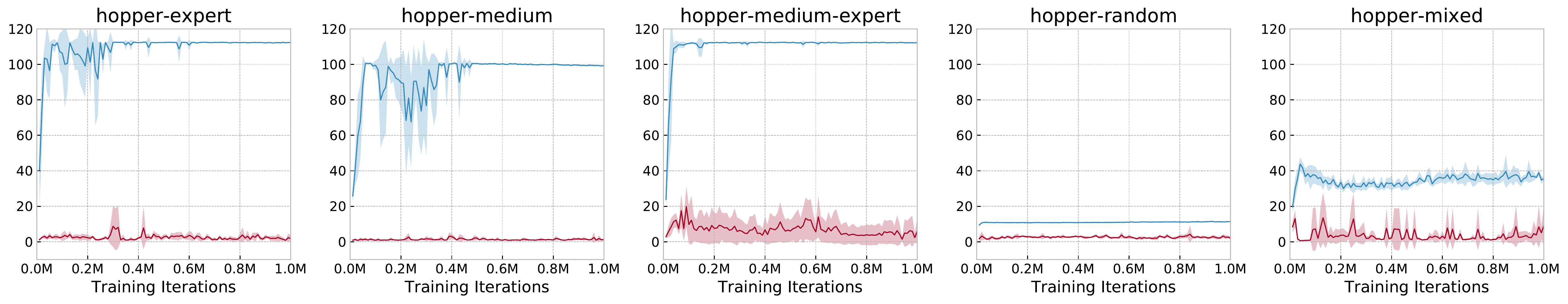} \\
    \includegraphics[width=\textwidth]{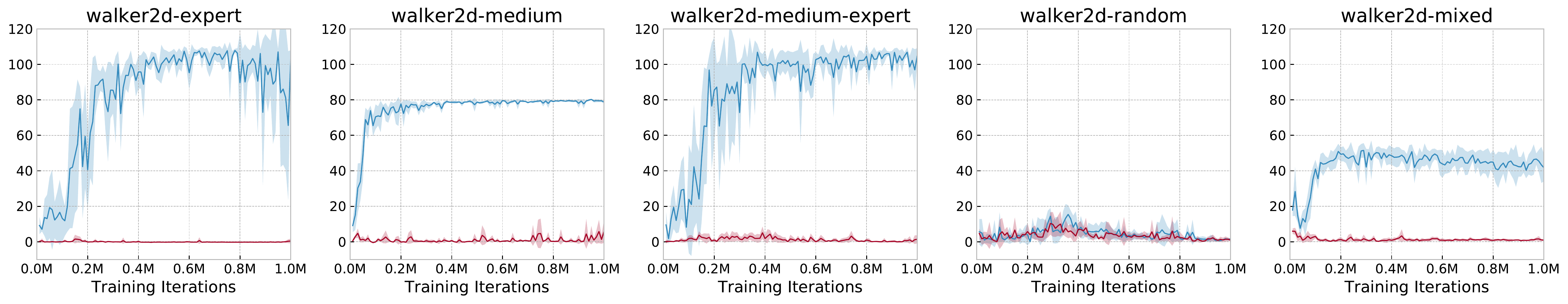}
\caption{Performance of F-BRC without our critic representation. Without the critic representation, the gradient penalty term alone fails to improve performance of the underlying reinforcement learning algorithm on the offline datasets. }
    \label{fig:abl_plots_2}
\end{figure*}

\end{document}